# A Brief Introduction to Temporality and Causality

Kamran Karimi
D-Wave Systems Inc. Burnaby, BC, Canada
kkarimi@dwavesys.com

**Abstract**
   Causality is a non-obvious concept that is often considered to be related to temporality. In this paper we present a number of past and present approaches to the definition of temporality and causality from philosophical, physical, and computational points of view. We note that time is an important ingredient in many relationships and phenomena. The topic is then divided into the two main areas of temporal discovery, which is concerned with finding relations that are stretched over time, and causal discovery, where a claim is made as to the causal influence of certain events on others. We present a number of computational tools used for attempting to automatically discover temporal and causal relations in data.

## 1. Introduction

   Time and causality are sources of mystery and sometimes treated as philosophical curiosities. However, there is much benefit in being able to automatically extract temporal and causal relations in practical fields such as Artificial Intelligence or Data Mining. Doing so requires a rigorous treatment of temporality and causality. In this paper we present causality from very different view points and list a number of methods for automatic extraction of temporal and causal relations.

   The arrow of time is a unidirectional part of the space-time continuum, as verified subjectively by the observation that we can remember the past, but not the future. In this regard time is very different from the spatial dimensions, because one can obviously move back and forth spatially. However, there is a contradiction between the empirical observation that time is irreversible, and the time-reversibility of physical laws. Most laws in physics are expressed in a time-reversible manner, meaning that they can be applied in either temporal direction. An example is the relationship between force and acceleration, $f = m \times a$ (1), which does not enforce a specific direction of time. While formula (1) might suggest that the force $f$ is caused by the mass and the acceleration, we can re-arrange the formula to read $a = f / m$ and interpret the formula as saying that acceleration is caused by the mass and the force. The fact that many physical laws do not exhibit temporal asymmetry has prompted some researchers to consider them as incomplete approximations [32]. Others assume that there are two distinct types of physical laws. *Time-symmetrical* laws hold backwards in time, while *asymmetrical* laws are valid in only one direction [3].

   Note that another possible view is that a formula such as $f = m \times a$ shows an instantaneous physical relationship: the force at any one moment is related to the acceleration at that same moment. In other words, they are created together and one cannot exist without the other.

A mug falling from the table and breaking is observed a lot more often than the reverse, and we always get older, but not younger. These irreversible phenomena are explained by the second law of thermodynamics, which states that the entropy of any closed system does not decrease over a sufficient long period of time [12]. In physics, entropy by definition makes time unidirectional, as we define the direction of time to be that of increasing entropy. Time being unidirectional is observed at the quantum level too, as evident by the quantum de-coherence, which is the irreversible collapse of a superposition of states to a single state [28]. Quantum de-coherence can be caused by a measurement.

Note that nothing forbids a broken mug to reassemble itself on a table, or for air molecules to all gather in a small part of the room, but the probability of such events happening is extremely small.

Causality is a hard to define concept. It has been studied in many different disciplines, including physics, philosophy, statistics, and computer science. Time has usually been an essential part of our intuitive understanding of causality, because we often consider it necessary that the cause *A* should have existed in time before the effect *B*.

Although causality is subject to debate, there is value is trying to discover possibly causal relations between events. If in a system under study we cannot change any variables (as in the weather), then understanding causal relations can help us predict the future by observing the value of current variable. If we can change certain variables, then knowing about casual relations will allow us to exert control on the behavior of the system.

In what follows we will discuss the problems of defining and discovering temporal and causal relations. We show that Computer Science has allowed us to go beyond philosophical discussions and attempt to discover temporal and causal relations with different degrees of success. The rest of the paper is organized as follows. In Section 2 we briefly discuss temporal data and mention different techniques of processing them. Section 3 provides different definitions of causality. Section 4 presents a number of algorithms for causal discovery from data, while Section 5 concludes the paper.

**2. Temporal Data and Discovery**

Temporal data are often represented as a sequence, sorted in a temporal order. Examples of studies of sequential data and sequential rules are given in [2, 10, 36]. There are a number of general fields in the study of sequential data. A *time series* is a time-ordered sequence of numerical observations taken over time [4, 5]. An example is the series of numbers <1, 3.5, 2, 1.7, …>. In a *univariate time series*, each observation consists of a value for a single variable, while in a *multivariate time series*, each observation consists of values for several variables.

Some approaches to time series processing have assumed the presence of a distinguished variable representing time, and numeric values for all other variables. Attempts have been made to fit constant or time-varying mathematical functions to time series data [4]. A time series can be *regular* or *irregular*, where in a regular time series data are collected at predefined intervals. An irregular time series does not have this property, and data can arrive at any time, with arbitrary temporal gaps in between. A *deterministic time series* can be predicted exactly, while the future values in a *stochastic time series* can only be determined probabilistically. The former is a characteristic of artificial and controlled systems, while the latter applies to many natural systems. Simple operations like determining the minimum or maximum values of certain variables, finding trends (such as increases or decreases in the value of stocks), cyclic patterns (such as seasonal changes in the price of commodities), and forecasting are common applications of time series data.

Many approaches to the discovery of rules from time series data involve pre-processing the input by extracting global or local features from the data. Global features include the average value or the maximum value, while local features include an upward or downward change, or a local maximum value [19]. Another example of discovering temporal traits by pre-processing time series data is the discovery of increasing or decreasing trends before rule extraction [17]. While the study of time series is pursued widely, there are hints that in some cases the results may not be useful or even meaningful [23].

In [29], the phrase *multiple streams of data* is used to describe simultaneous observations of a set of variables. The streams of data may come from different sensors of a robot, or the monitors in an intensive care united, for example. The values coming out of the streams are recorded at the same time, and form a time series. In [26] an algorithm is presented that can find rules, called "structures" by the authors, relating the previous observations to the future observations. Such temporal data appear in many application areas and a good review can be found in [34].

An *event sequence* is a series of temporally ordered events, with either an ordinal time variable, which gives the order but not a real-valued time, or no time variable. The main difference between an event sequence and a time series is that a time series is a sequence of real numbers, while an event sequence can contain variables with symbolic domains. Each *event* specifies the values for a set of variables. A recurring pattern in an event sequence is called a *frequent episode* [25]. Recent research has emphasized finding frequent episodes with varying number of events between key events that identify the event sequence. Algorithms such as *Dynamic Time Warping* and its variants measure the similarity of patterns that are stretched differently over time [21].

Temporal sequences are often considered to be passive indicators for the presence of temporal structure in data [1, 11, 25, 29], and no claim is made as to whether or not they represent causal relationships. Despite having different terminology, all the domains listed so far in this section have the common characteristic of recording the values of some variables and placing them together in a record. Time series, event sequences, and streams of data all try to find temporal rules, called patterns, episodes, and structures, respectively, from the input data. An extensive review of the methods of discovering knowledge from sequential data can be found in [16].

## 3. The Elusive Causality

Causality has been studied in many different disciplines. Philosophy's look at causality has changed greatly over time, reflecting the scientific understanding of the era. Hume's notion of causality is very close to our common sense understanding of this concept. It states three conditions for causality: *A* causes *B* if 1) *A* precedes *B* in time, 2) *A* and *B* are contiguous in space and time, and 3) *A* and *B* always co-occur or neither of them occurs [15]. The main problem with this definition is that one can argue that day is causing night, or vice versa, as all three conditions are satisfied in this case.

Mill's definition of causality also has three conditions: 1) temporal precedence, 2) association of the cause and the effect, and 3) absence of other plausible causes [15]. The last criterion fixes the problem of day and night. Rubin believes that causality should be limited to specific contexts and conditions. According to him, any relationship that is derived from experimentation, where certain variables are changed in an experimental group and fixed in a control group, is a causal relation [15]. This idea has been the basis of many empirical studies in sciences such as medicine.

Causality is of importance in areas such as religion and law too. In some religions a causeless creator is considered the ultimate cause for everything. In many such religions people can still cause events and will be punished or rewarded for them based on the effects and the intentions behind them. Other than the "heavenly court," believed by followers of certain religions, there is an apparent need to assign *responsibility* for certain events in the every-day life. For this reason, Lawyers often need to establish a *sufficient causal link* before they can hold a person or other entity responsible. Lawyers assume that an entity has caused an unlawful event if an action has been performed by the entity that could lead to the event, and without that action, the event would not have happened.

People usually consider it necessary that the cause *A* should have existed in time before the effect *B*. However, modern physics is not based on intuitive ideas about time. There is no universal clock, and the only constant is the speed of light. Each observer perceives time differently according to his speed relative to another observer. Physicists consider the speed of light to be an upper limit for the speed with which an event *A* can cause another event *B* [12].

There are theories about particles that can move faster than light. One example is the tachyon [7]. But in the macroscopic world, moving faster than light may lead to contradictions, as the effect *B* can appear before the cause *A*. This possibility can lead to paradoxes: if a person can move faster than light, then he can return to his past and change the causes of his travel in time. The effect happening before the cause is called *backward causation*. As shown in [42], some philosophers do not consider this impossible or paradoxical. It is possible that in the absence of free will, we could go back to our past and not change anything that interferes with our time travel. An implication of this assumption is that since we are not able to change anything, we would travel to the past again and again, and thus be stuck in a never-ending time loop.

In classical Newtonian physics, causality is well-defined. As exemplified by a claim by Laplace, it was believed that if one knows the initial states of all the particles in the Universe, plus the applicable rules, one can predict the future or retrodict the past perfectly [28]. More specifically, given the current position value (*x, y, z*), the current momentum vector ***p***, plus all the acting forces, in classical physics one can plot the trajectory of the particle and predict the future position and momentum [28].

Quantum theory, however, has placed severe restrictions on predictability in the classical physics sense. A consequence of the Heisenberg uncertainty principle is that one cannot know both the position and the momentum of a particle with arbitrary precision. More precisely, $\Delta x \times \Delta \mathbf{p} > 0$, where the $\Delta$ operator denotes the uncertainty in the value of its operand [28]. There is another example of how quantum physics is in disagreement with the intuitive understanding of causality: while in classical physics the interaction of a cause and its effect requires spatial proximity, quantum physics allows apparently instantaneous causal interactions between particles that are at arbitrary distances from each other. As an example, given a set of two entangled electrons, measuring the spin direction of one of them instantly determines the spin direction of the other one, thus establishing causal non-locality [24].

Granger causality, used mainly in Economics, puts clear emphasis on temporal precedence [9]. A *Granger causal relationship* exists when previous values of some variables improve the prediction of a decision variable's value. Suppose we are observing two variables $x_t$ and $y_t$ over time, and $A$ is a set of variables considered relevant to $y_t$. We say $x_t$ granger-causes $y_t$ if there is a natural number $h > 0$ such that $P(y_{t+h} | A) < P(y_{t+h} | A \cup \{x_t, x_{t-1}, …\})$, where $P(a | b)$ measures the conditional probability of *a* given *b*, i.e., the probability of event *a* happening, given that event *b* has already happened.

For example, knowing the previous value of the variable $x_t$ = "Was there a political scandal today?" during the past few days may increase our ability to predict the value of the variable $y_t$ = "Does the stock market lose value today?". Granger causality is subject to errors. For example, if the event "person *P* leaves the building" is always followed 10 minutes later by "Everybody leaves the building" then Granger causality will consider person *P*'s leaving the building as a cause for everybody leaving the building, which may or may not be true.

A major trend in the development of the concepts of causality is that they have become more computationally verifiable with time. In a complex world, providing clear-cut definitions and measures for causality is hard, and this consideration is present in more recent statistical approaches such as Granger causality or conditional dependence (and independence), which have been implemented on computers.

Probabilistic causality can be defined as follows. *A* is considered to be a cause of *B* if we have $P(B | A) = 1$ and $P(B | \sim A) = 0$. This formula implies temporal precedence of the cause with regards to the effect. In practice, however, this definition is too brittle, because of possible errors in gathering data for example, and few real-world data would satisfy it. An attempt to remedy this brittleness is to assume that *A* is a cause of *B* if $P(B | A) > P(B | \sim A)$. An example of a weakness of this second definition comes into view if *A* = seeing lightning, and *B* = hearing thunder. Suppose we hear the thunder only after seeing the lightning, then we have $P(B | A) > P(B | \sim A)$, which would lead us to believe that the act of seeing the lightning causes the hearing of thunder. In this example, the lighting and the thunder are created at the same time by a discharge of electricity. For this reason in the probabilistic approach there is an added assumption that there should not be any hidden common causes at work.

In our description of the statement P(*B* | *A*), event *A* is considered to have already happened, but this consideration does not enforce an automatic temporal ordering, because according to Bayesian rule: P(*B* | *A*) = P(*A* | *B*) × P(*B*) / P(*A*). In the left hand side of the equation event *B* is assumed to have already happened, while in the right hand side event A is assumed to have happened. In other words, by using algebraic manipulation we can change the order in which the events are supposed to have happened, and thus reverse the original, and possibly natural, order of the events.

When a conditional probability value expresses a reverse temporal ordering, then it is called the *likelihood*. Suppose we create a model *M* from some data *D* and then measure the conditional probability of the data given the model P(*D* | *M*). Since the data existed before the model, this probability is called the likelihood of *D* given *M*, and indicates reversing the temporal order because now the model explains the data, instead of the other way around. For more discussions about discovering causality, especially from a statistical point of view, refer to [8, 18, 22, 40, 45].

**4. Causal Discovery in Computer Science**

Automatic discovery of causal relations among a number of variables has been an active research field. More specifically, automated methods have been applied to determining whether or not the value of a variable is caused by other variables.

The input to a computational approach usually consists of data records, each containing the values of variables observed together. The Bayesian approach [13] to the definition and discovery of causality is currently enjoying much attention, and is introduced in Section 4.1. We also briefly present a *Minimum Message Length* method [43] of causal discovery in Section 4.2. None of these two methods take any temporal information into consideration, so we introduce an approach that works based on explicit temporal ordering among variables in Section 4.3.

**4.1 The Bayesian Approach**
A prevalent approach to the discovery of causality is to consider the problem to be that of creating a graph, where the parent nodes denote causes, while the children denote effects. One can argue that the variables put in a graph form mere temporal associations, so interpreting a relation as causal requires justification. The main trend in causality mining involves using the statistical concept of conditional independence as a measure of the control one variable may have over another [31]. For example, given the three variables *x*, *y*, and *z*, if *y* is independent from *z* given *x*, that is, $P(y, z | x) = P(y | x)$, then we can conclude that *y* is not a direct cause of *z*. In other words, *x* separates *y* and *z* from each other. This basic concept is used to build *Bayesian Networks*, which show the conditional dependence of variables as arcs [14, 30]. These arcs are then interpreted as signifying causal relations. Bayesian networks are thus directed acyclic graphs that represent the conditional dependency of the variables.

TETRAD [37, 38] is a well-known application for causal discovery that uses Bayesian networks for attempting to find causal relations. A *causal* Bayesian network, as used in TETRAD, assumes that the links in the graph denote causal relationships. TETRAD at first assumes that all variables are causally related, so at the beginning the causal network is fully connected. Then it uses conditional independence tests to remove or revise edges. The remaining edges form a causal Bayesian network.

In constructing a Bayesian network, we simplify joint probability distributions by using products of marginal probability distributions and conditional probabilities. As a simple example, P(*a*, *b*) can be written as P(*b* | *a*) × P(*a*). In this case the corresponding Bayesian network is a graph consisting of two nodes *a* and *b*, where *a* is the parent of *b* (node *a* points to node *b*). In this case event *a* is a parent of event *b*, and there is a directed link from node *a* to node *b*. In general, the chain rule in probability is written as $P(x_1, \ldots, x_n) = P(x_n | x_{n-1}, \ldots, x_1) \times P(x_{n-1} | x_{n-1}, \ldots, x_1) \times \ldots \times P(x_2 | x_1) \times P(x_1)$. This formula can be simplified when we have information about the conditional independence of the variables, which can be either given by the user, or computed from the data. Figure 1 shows an example Bayesian network corresponding to simplifying P(*f*, *t*, *c*, *a*, *r*) = P(*r* | *f*, *t*, *c*, *a*) × P(*a*, | *f*, *t*, *c*) × P(*c* | *f*, *t*) × P(*t* | *f*) × P(*f*) to P(*f*, *t*, *c*, *a*, *r*) = P(*r* | *a*, *c*) × P(*a* | *f*, *t*) × P(*t*) × P(*f*) × P(*c*).

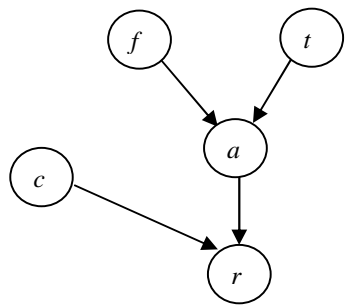

**Figure 2.1** An example Bayesian network.

This Bayesian network implies the property that any node is independent of its non-descendant nodes when conditioned on its parents. This implication is called the *causal Markov condition* [41]. Similar graphs appear in belief networks, probabilistic networks, knowledge maps, and causal networks [35].

We should emphasize that the notion of conditional independence as defined in statistics is devoid of time. The proponents of this method use temporal information, if available, to place constraints on the relationships among the variables (if we know *x* always happens before *y*, then *y* cannot be a cause of *x*), but time is not essential to the working of such algorithms.

A practical problem with Bayesian networks is that constructing a network from observed data is an NP-Hard task [6], and becomes increasingly time-consuming, or impossible, as the number and combination of variables increase.

### 4.2 The Minimum Message Length Method

MML is introduced in [43] and is based on maximizing the posterior probability of a model. If we want a good model $M$ which is fitted to the data $D$, then we want to maximize the posterior probability $P(M \mid D)$. According to Bayes' rule we should maximize $P(D \mid M) \times P(M) / P(D)$. We recall that $P(D \mid M)$ is the likelihood of $D$ given $M$. Since $D$ is given, we consider $P(D)$ a constant, so the model $M$ must be chosen such that the numerator is maximized. One can employ the information theory's [39] method of taking the negative of the natural logarithm of the probability values, and convert the problem to that of minimizing the expression $-\ln(P(D \mid M)) - \ln(P(M))$. In information theory, this formula expresses the minimum length necessary to encode the model $M$, which is $-\ln(P(M))$, plus the minimum length necessary for encoding any exceptions to the rules in the data, expressed by $-\ln(P(D \mid M))$. Hence the name MML.

CaMML (Causal Minimum Message Length) is an application that attempts to learn the best causal Bayesian structure to explain some observed data, using an MML metric for selecting a causal Bayesian network. As explained above it measures the goodness-of-fit of a causal model to the data [44]. Given a set of observed variables, CaMML finds causal relationships between one or more causes and a single effect.

CaMML searches the space of possible causal models using the Markov Chain Monte Carlo Method (MCMC) [27], and finds the one that best explains the data. In a Markov process, the transition from one state to the next depends on the current state only. In other words, any transition in a Markov process does not depends on the history of the moves that lead to the current state. A Markov chain is characterized by a transition matrix that gives the probability of moving from one state of the system to the next one. Starting from an initial state and multiplying by the Markov transition matrix enough number of times, we settle in a final state. Monte Carlo methods work by simulating an unknown function using probabilistic means. They sample values from a probability distribution and compute a function at those points. In MCMC, to obtain a specific probability distribution, one generates a Markov chain whose long-term equilibrium is that distribution.

### 4.3 Discovering Causality from Sequential Observations

The Temporal Investigation Method for Enregistered Record Sequences (TIMERS) method takes a classification approach to the discovery of causality [20]. The input to this method is considered to consist of a sequence of records, each containing the values of a number of variables. The records are observed one after the other. An example would be the two record sequences <1, 2, true> and <2, 3, false>, where the values of three variables $<v_1, v_2, v_3>$ are noted. The time interval between registering the records is determined by the problem domain. Assuming that the effects take time to manifest, TIMERS merges consequent records together. The number of records that are merged together is determined by a window size.

Merging allows the potential causes and effects to appear in the same record, making it possible to use a time-agnostic learning algorithm to be used for causal discovery. Using a window size of 2, merging the above two records would result in: <1, 2, true, 2, 3, false>, which corresponds to the values of variables <$v_{11}$, $v_{12}$, $v_{13}$, $v_{21}$, $v_{22}$, $v_{23}$>. We have added a subscript to denote the temporal order of the variables and so help distinguish the same variable as it appears at a different time. After that one generates decision rules from the merged records using conventional classifiers such as C4.5 [33].

To find causality, one intends to find rules to predict the current value of a single variable using the past values of other variables. After generating decision rules from the merged records, the algorithm notes the quality of the rules using their training or predictive accuracies. The success or failure of the attempt to explain the value of the decision variable using the previous values depends on the quality of the rules: If the training or predictive accuracy values are above a user-defined threshold, then the explanation is considered to be satisfactory, and the target variable's value is said to have been caused by the values of other variables used in the decision rule.

In the above example previous observations were used to explain the current observation, which is how TIMERS defines causality. It considers two other possibilities: a relation being instantaneous, or being acausal. In the first case simultaneous observations are used to make a prediction, while in the second case future observations are used to predict the current value. The decision on the type of the relationship depends of the accuracy of the resulting rule sets. For example, if an acausal set of rules results in better accuracy than instantaneous and causal rules, then an acausal relationship is said to be at work.

An acausal relationship can be considered to be the same as a backward causal one, where the future is supposed to have caused the present. One can consider acausality to denote the presence of hidden common causes in the past. In other words, in TIMERS an acausal relation is considered to imply a temporal co-occurrence, where some events are happening over time and form a temporal pattern, but none is causing the other.

## 5. Concluding Remarks

In this paper we briefly presented some of the many different approaches to defining and discovering temporal and causal relations in a number of fields, with an emphasis on computational methods. None of the presented method can claim to have solved the problem of reliable causal discovery, and the results should be subject to verification.